# Around View Monitoring System for Hydraulic Excavators

Dong Jun Yeom[1], Yu Na Hong[1], Yoo Jun Kim[1], Hyun Seok Yoo[2], Young Suk Kim[1*]

[1] *Department of Architectural Engineering, Inha University, Incheon 22212, Korea*
[2] *Department of Technology Education, Korea National University of Education, Cheongju 28173, Korea*
E-mail address: youngsuk@inha.ac.kr

**Abstract:** This paper describes the Around View Monitoring (AVM) system for hydraulic excavators that prevents the safety accidents caused by blind spots and increases the operational efficiency. To verify the developed system, experiments were conducted with its prototype. The experimental results demonstrate its applicability in the field with the following values: 7m of a visual range, 15fps of image refresh rate, 300ms of working information data reception rate, and 300ms of surface condition data reception rate.

**Key words:** Safety, Operational Efficiency, Around View Monitoring, Hydraulic Excavator

## 1. INTRODUCTION

Blind spots on construction equipment resulting in operator's poor visibility are one of the leading causes of safety accidents in the construction industry [1-4]. They create problems for operators by restricting their line-of-sight and impeding their view of personnel and materials in an operational area [3, 4]. According to the analysis of industrial accidents in the Ministry of Employment and Labor in Korea, fatal accidents occurred by construction equipment accounted for 27.5% of total fatal accidents of the construction industry in 2015. Its death toll has remained over 83 from 2010 to 2015 [5].

Applying Around View Monitoring (AVM) system to construction equipment can be an alternative to prevent the related safety accidents. This system is a technology that provides a top-view image of the automobile's surroundings in real time through cameras installed on each side of it. It was originally developed for assisting commercial automobiles when parking and driving. Since the technology enables an operator to see the surroundings, this allows people to drive an automobile with increased convenience, efficiency, and safety [6-10]. With its usefulness, the interest in applying AVM systems to hydraulic excavators, the representative construction equipment, has continuously increased in domestic and foreign construction industries [11-13].

However, the existing AVM systems for hydraulic excavators were developed based on the same technology for commercial automobiles. They have practical limitations due to an insufficient visual range of display for excavation work and inflexibility to reflect excavators' features into the systems, such as movements of excavator's boom, arm, and bucket. Moreover, they are only available on flat ground as severe distortion on display when an excavator is tilting [14, 15].

In this study, the authors developed the AVM system that is available on undulating grounds and that empowers an excavator operator to obtain working information in real time with its sufficient visual range. This system aims at improvement of safety and operational efficiency in excavation work.

## 2. AVM SYSTEM FOR HYDRAULIC EXCAVATORS

The AVM system for hydraulic excavators in this study was developed with three core technologies: i) AVM display technology, ii) representing technology for working information, iii) calibration technology for a change of excavator's inclination, as shown in Fig. 1. These three core technologies were defined through literature reviews, on-site surveys, and interviews with excavator operators.

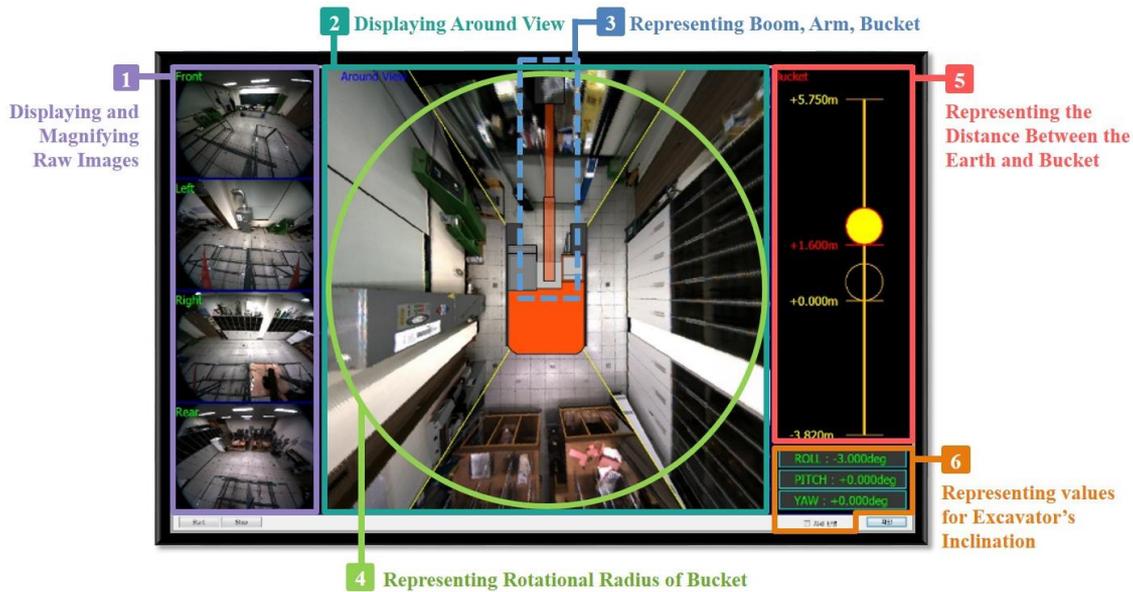

**Fig. 1**. Prototype of AVM system for hydraulic excavators

### 2.1. AVM display technology

Considerable safety accidents such as injuries and fatalities do occur because of operator's poor visibility of personnel and materials on blind spots [1-4]. The primary purpose of this study is to alleviate these safety problems by eliminating the blind spots.

AVM display technology is a system that provides an operator a 360-degree top-view image of excavator's surroundings and raw images in each of four cameras (Fig. 2). Its visual range is set to display a full-view of excavator's working radius and its raw images on the left side of the screen that can be magnified to a full-scale image if necessary. This technology is beneficial to operators in order to improve safety and efficiency in excavation work by monitoring a top-view image for the operational area, and raw images in real time.

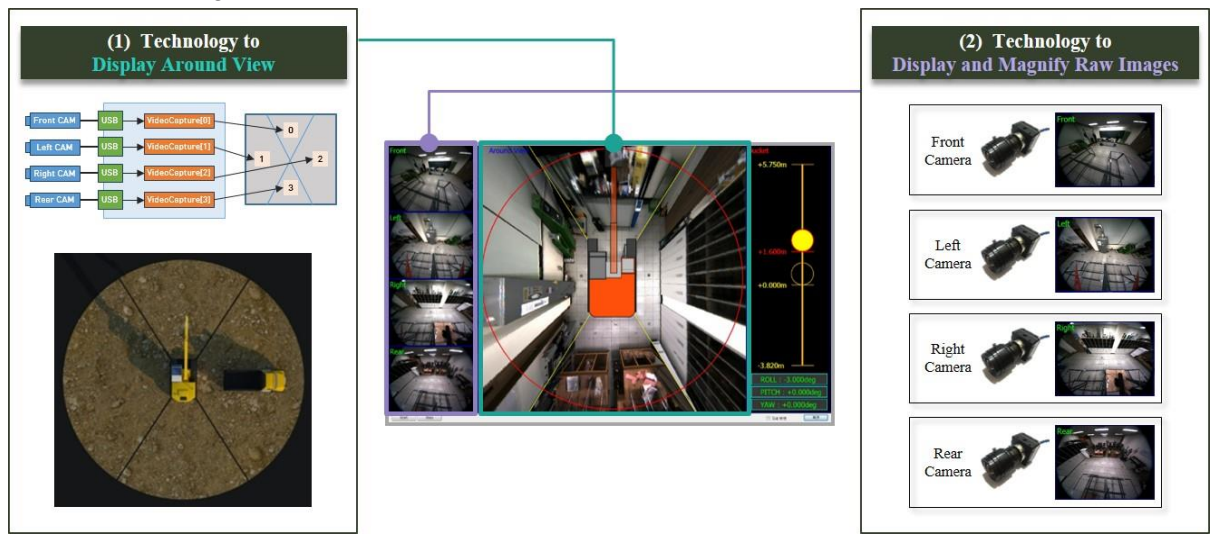

**Fig. 2**. AVM display technology

### 2.2. Representing technology for working information

Operating a hydraulic excavator can be a cumbersome undertaking because its Human-Machine Interface (HMI) is not easy to use. Becoming a skilled operator of excavator requires years of rigid training, hands-on practice, and education [16-18]. This operational difficulty would increase operating errors and the time needed to learn the operation for inexperienced operators [19].

Representing technology for working information is a system that visually illustrates the boom, arm, bucket, rotational radius of the bucket, and the distance between the surface of the earth and the bucket, as shown in Fig. 3. With this technology, the operation becomes easier, faster and more effective that can ameliorate its efficiency, convenience, and accuracy in excavation work.

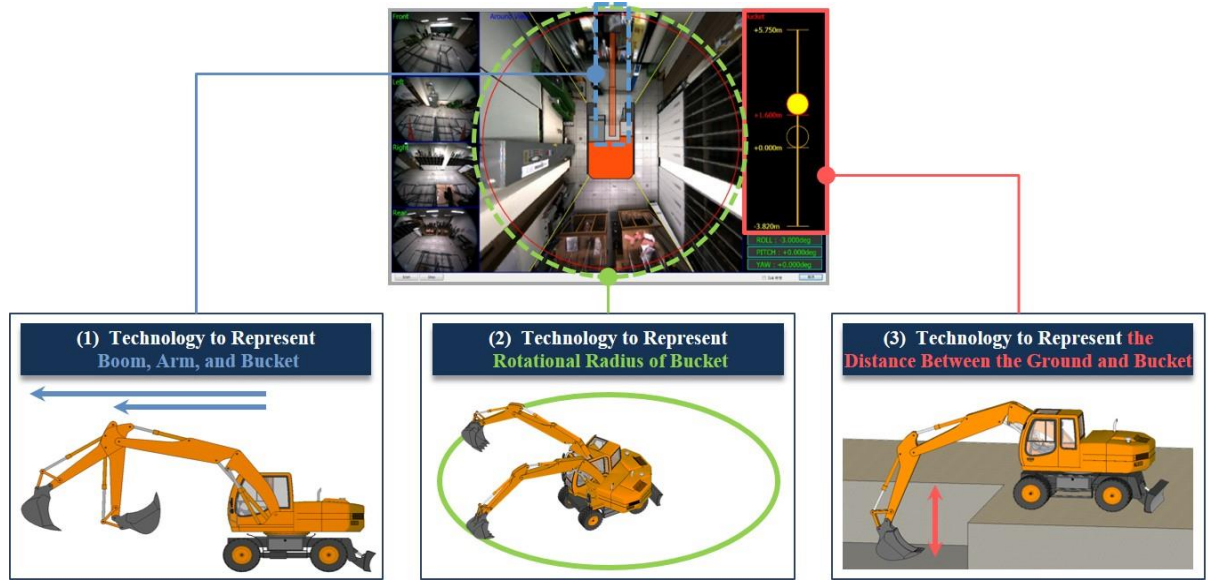

**Fig. 3.** Representing technology for working information

## 2.3. Calibration technology for a change of excavator's inclination

Hydraulic excavators are often used on slopes and uneven grounds that can result in safety accidents when the angle of inclination in the operational area is greater than its acceptable value [20-22]. Accordingly, the excavator's inclination creates a distortion on AVM display. Fig. 4-(2) depicts the change of image scope according to a change of an excavator's inclination [14, 15]. This change of the image scope distorts the top-view image, as it is a stitched image of the four separate images from the cameras [10]. For the sake of safety and practicality, therefore, the AVM system should cope with the problems caused by excavator's inclination.

Calibration technology for a change of excavator's inclination is a system that represents values of roll, pitch, and yaw for an excavator's position, and calibrates the distortion caused by an excavator's inclination (Fig. 4). This technology is essential to operators to the extent of using the AVM system on undulating grounds and improving safety in excavation work by acquiring an excavator's positional data.

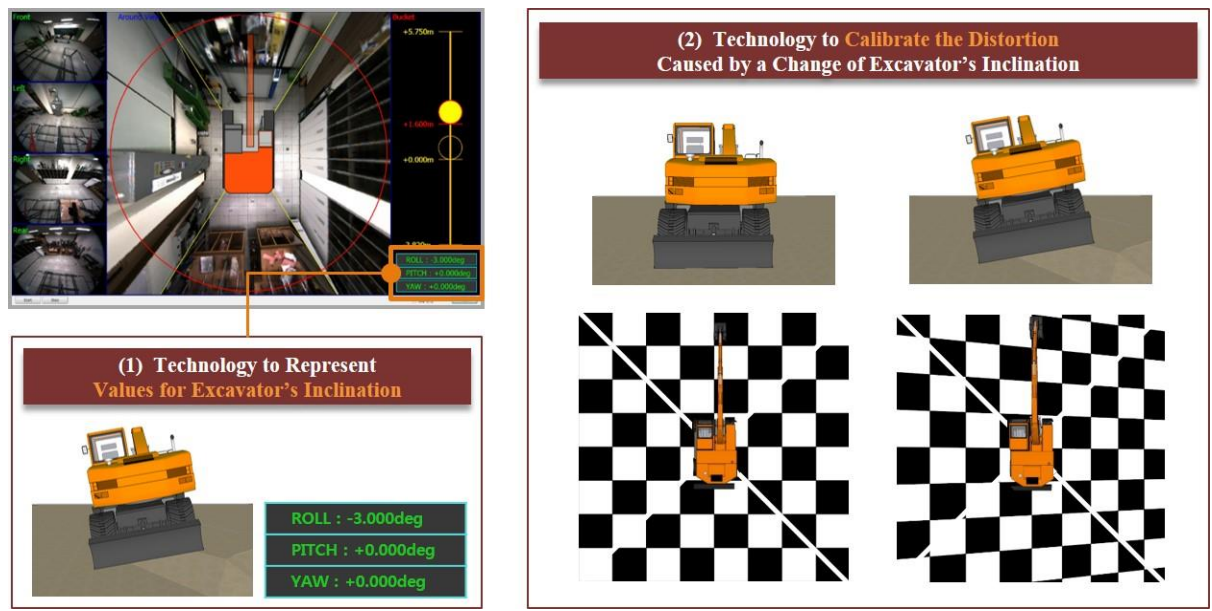

**Fig. 4**. Calibration technology for a change of excavator's inclination

## 3. DESIGN OF AVM SYSTEM FOR HYDRAULIC EXCAVATORS

### 3.1. Relation analysis to select proper cameras and lens

To eliminate the blind spots in excavation work, the display range of AVM system should include a hydraulic excavator's maximum working radius. In this regard, it is crucial to select cameras and lens having an enough field of view for the display range. Therefore, we analyzed the relations among display range, image range, a field of view, focal length, and size of the image sensor shown as follows (Fig. 5).

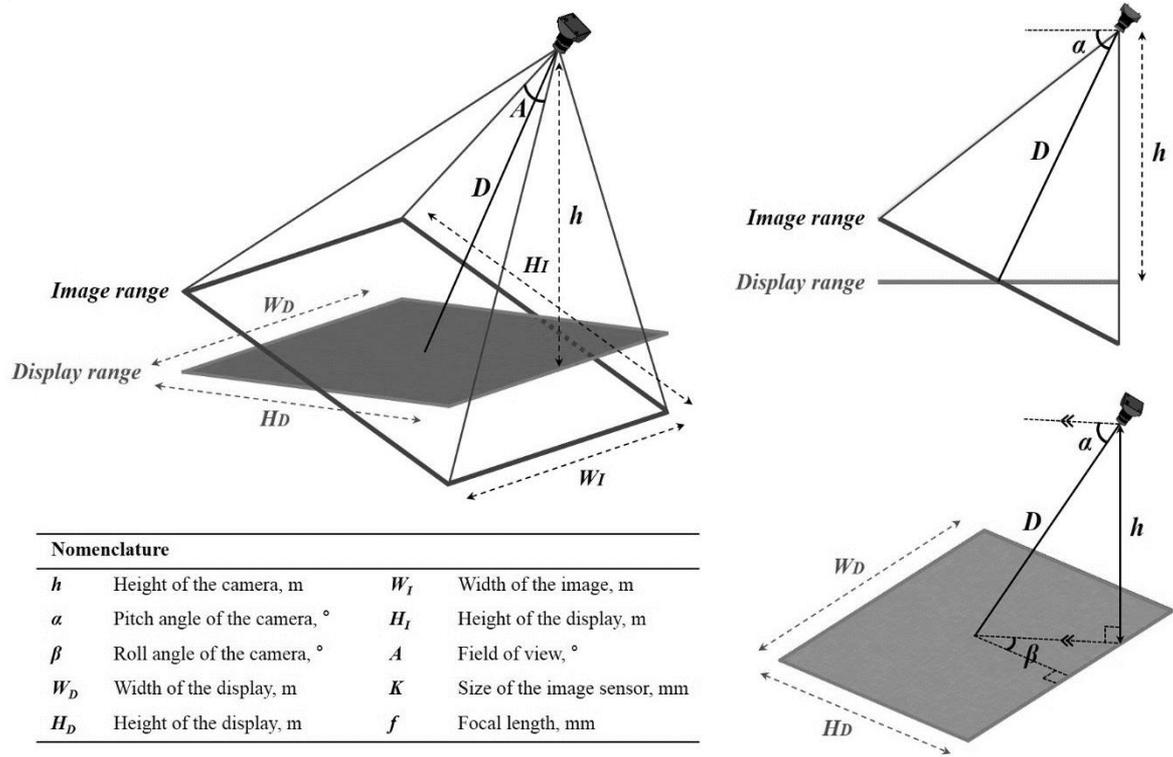

**Fig. 5**. Geometry of a camera and its nomenclature

When camera positions and the display range are determined, the image range is given by:

$$W_I = (W_D \times \cos\beta) + (H_D \times \sin\beta) \tag{1}$$

$$H_I = \{(W_D \times \sin\beta) + (H_D \times \cos\beta)\} / \sin\alpha \tag{2}$$

The field of view may then be obtained since:

$$A = 2\tan^{-1}(\sqrt{W_I^2 + H_I^2} / 2D) \tag{3}$$

Also, the field of view can be found as [23]:

$$A = 2\tan^{-1}(K/2f) \tag{4}$$

where $K$ is the size of the image sensor and $f$ is the focal length. The equation also is represented as:

$$K/f = 2\tan(A/2) \tag{5}$$

When the display range is determined, therefore, the authors are able to calculate a proper field of view and a value of $K/f$.

The excavator for the prototype in this study is the widely used excavator in Korea, Doosan DX140W hydraulic excavator. Its dimension is 2.5m in width, 3.5m in depth, and 2.2m in height. The excavator's maximum working radius is 7.8m. Fig. 6 and Table 1 illustrate the each camera positions and the display range.

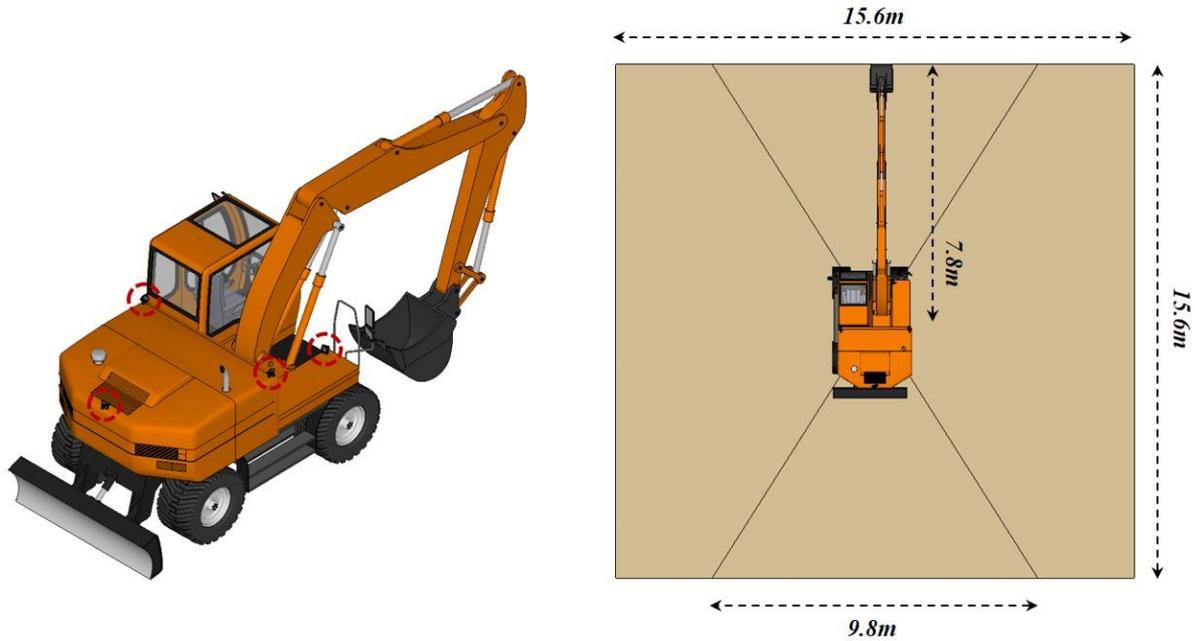

**Fig. 6**. Camera positions and the display range

**Table 1.** Values of each camera positions and its display range

|  | **h(m)** | **α (°)** | **β (°)** | **W$_D$ (m)** | **H$_D$ (m)** | **D (m)** |
|---|---|---|---|---|---|---|
| Front camera | 1.50 | 24.40 | 10.80 | 9.80 | 6.50 | 3.63 |
| Side cameras | 2.20 | 33.90 | 0.00 | 15.60 | 6.55 | 3.94 |
| Rear camera | 2.20 | 37.90 | 0.00 | 9.80 | 5.65 | 3.58 |

Therefore, the minimum value of $K/f$ for each camera is deducted as front camera 6.24, side cameras 4.96, rear camera 3.75 as shown in Table 2.

**Table 2.** Minimum value of $K/f$

| **Front camera** | **Side cameras** | **Rear camera** |
|---|---|---|
| $6.24 < K/f$ | $4.96 < K/f$ | $3.75 < K/f$ |

### 3.2. Coordinate geometry of a hydraulic excavator

The technology defined in Section 2.2 is a system that enables an excavator operator to acquire the positional information: (i) boom, arm, bucket, (ii) rotational radius of the bucket, and (iii) the distance between the surface of the earth and the bucket. To represent the positional information on the AVM screen, the geometry logic should be analyzed to convert the angular data received by sensors to the coordinate data of the positional data.

Therefore, we developed the forward kinematic equations for the positional information based on the precedent study [24]. Fig. 7 and Equation. (6)-(9) are the results of the coordinate geometry. The (0, 0) reference point for the coordinate system is the point of attachment of the excavator boom to the excavator with positive motion in the x-direction being away from it, and positive motion in the y-direction being vertically upwards [25]. $\theta_{bm}$ and $\theta_{arm}$ are the angles measured by sensors on the basis of a plane vertical to the gravitational acceleration, and $\theta_{bkt}$ is the angle gauged by sensors from a plane parallel to the gravitational acceleration [24].

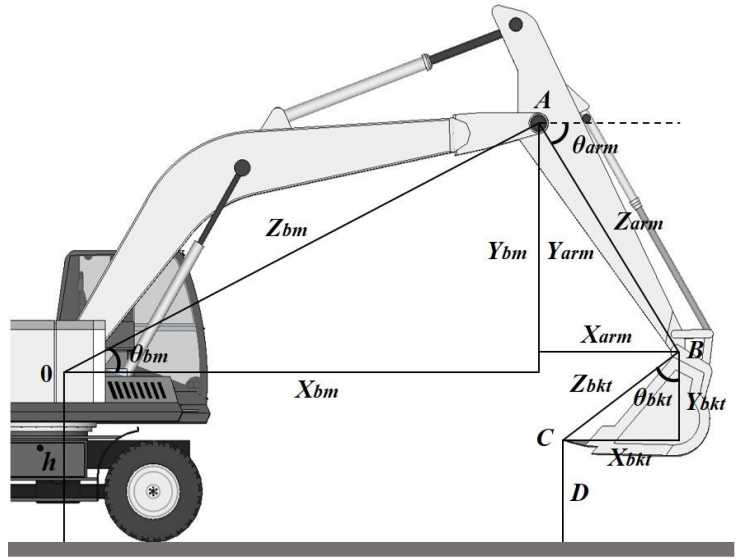

**Fig. 7**. Coordinate geometry of a hydraulic excavator and its nomenclature [24]

$$A = (X_{bm}, Y_{bm}) = (Z_{bm} \cos \theta_{bm}, Z_{bm} \sin \theta_{bm}) \quad (6)$$

$$B = (X_{bm} + X_{arm}, Y_{bm} - Y_{arm})$$
$$= (Z_{bm} \cos \theta_{bm} + Z_{bm} \cos \theta_{bm}, Z_{bm} \sin \theta_{bm} + Z_{bm} \sin \theta_{bm}) \quad (7)$$

$$C = (X_{bm} + X_{arm} - X_{bkt}, Y_{bm} - Y_{arm} - Y_{bkt})$$
$$= (Z_{bm} \cos \theta_{bm} + Z_{bm} \cos \theta_{bm} - Z_{bkt} \sin \theta_{bkt}, Z_{bm} \sin \theta_{bm} + Z_{bm} \sin \theta_{bm} - Z_{bkt} \cos \theta_{bkt}) \quad (8)$$

$$D = Y_{bm} - Y_{arm} - Y_{bkt} + h = Z_{bm} \sin \theta_{bm} + Z_{bm} \sin \theta_{bm} - Z_{bkt} \cos \theta_{bkt} + h \quad (9)$$

## 4. EXPERIMENTAL VERIFICATION

To verify the AVM system for hydraulic excavators developed, the authors conducted laboratory experiments with its prototype. The operating system of the prototype was Microsoft Windows 7, and Microsoft Visual Studio 2013 C++ and open CV were used as a software development environment. Its cameras and lens, u-Nova 20M/C and SY125M respectively, have the capacity to shoot pictures at 1600*1200 pixels resolution and its field of view is 148 degree. Fig. 8 shows the testbed that its size is equivalent to the size of Doosan DX140W hydraulic excavator.

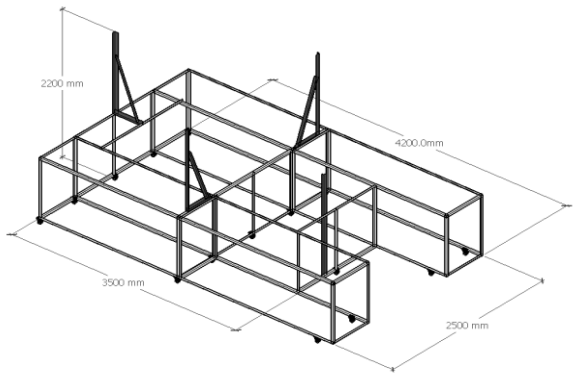 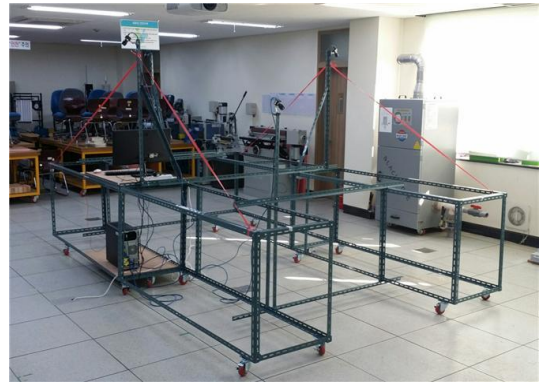

**Fig. 8**. Testbed for experiment

Fig. 9 shows the experimental results that the prototype displays up to 7m of visual range at over 15fps of refresh rate. The value of refresh rate is sufficient to display moving images either when materials proceed in the operational area or when the testbed moves back and forth – both can transpose simultaneously [26, 27]. In addition, raw images in each of four cameras are displayed with the same refresh rate and can be magnified to a full-scale image when a user touches it.

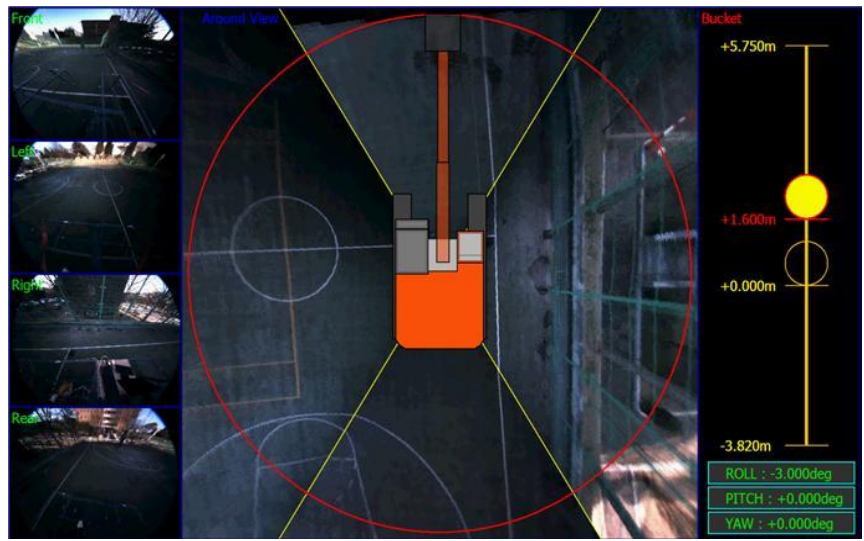

**Fig. 9**. Verification of AVM display technology

With the 300ms of working information data reception rate, it was verified that the prototype enables to display working information data such as boom, arm, bucket, rotational radius of the bucket, and the distances between the surface of the earth and the bucket in real time. Since it is a laboratory experiment with initial prototype before field-tests, the working information data used is not the actual data from sensors but the artificial data created for the experiment (Fig. 10).

Lastly, Fig. 11 demonstrates the experimental results of the calibration technology for a change of excavator's inclination. The Fig. 11-(1) represents the distortion caused when the roll angle of the excavator is at 3 degrees before it applies the calibration technology. On the other hand, Fig. 11-(2) is the resulting image that the prototype calibrated the distortion according to the excavator's positional data.

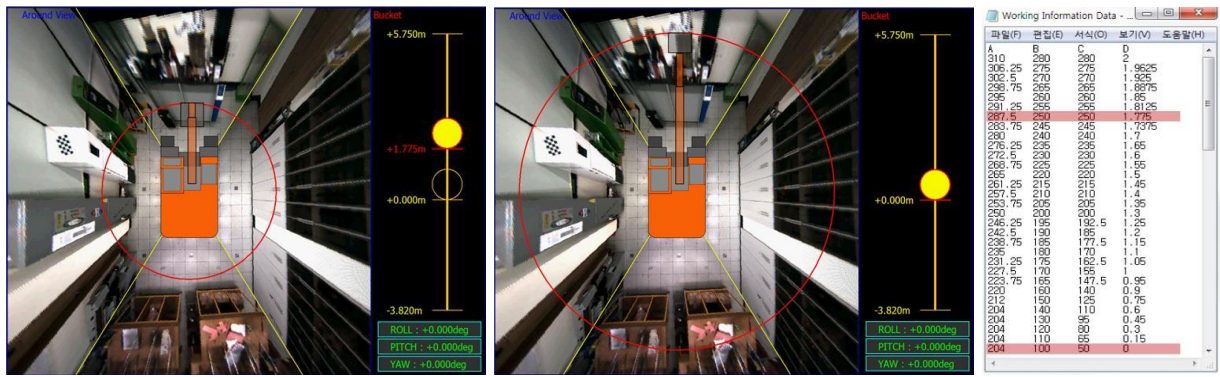

**Fig. 10**. Verification of representing technology for working information

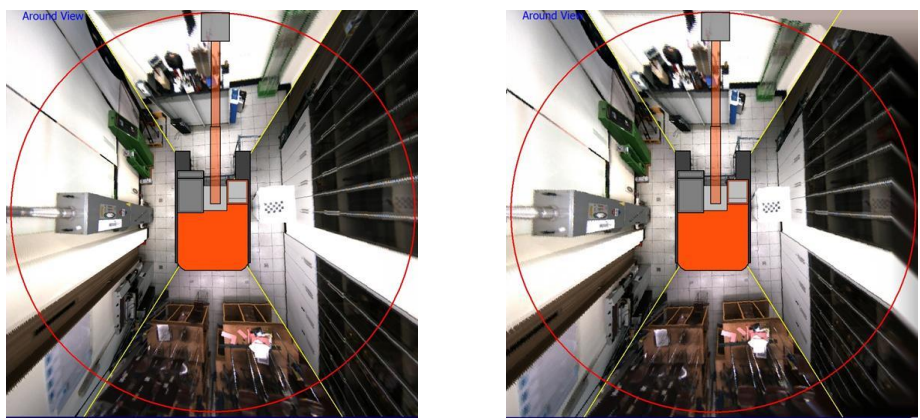

(1) Before the calibaration    (2) After the calibration

**Fig. 11**. Verification of calibration technology for a change of excavator's inclination

## 5. CONCLUSIONS AND FUTURE RESEARCH

This study researched the needs of the AVM system for hydraulic excavators in the field, its technical requirements and development, and its specifications. As an outcome, it was verified with the prototype that it displays 7m of an excavator's surroundings, represents working information, and calibrates the distortion caused by an excavator's inclination.

The scope of this research is limited to develop the AVM system and conduct laboratory experiments with its initial prototype. Therefore, the future study should carry out manifold actual field-tests and reflect the testing results into the system for its better functionality and applicability. It is envisioned that the ultimate AVM system for hydraulic excavators will play a vital role for their safer and more expeditious operations in the construction industry.

## ACKNOWLEDGEMENTS

This work is part of research conducted in the construction engineering and project management laboratory at Inha University. This research was supported by a grant(14SCIP-B079344-01) from Infrastructure and transportation technology promotion research program funded by Ministry of Land, Infrastructure and Transport of Korean government.

## REFERENCES

1. Pratt G, Fosbroke E, Marsh M, 2001. Building safer highway work zones: Measures to prevent worker injuries from vehicles and equipment. Department of Health and Human Services, CDC, NIOSH, 1.

2. Ray SJ, Teizer J, 2013. Computing 3D blind spots of construction equipment: Implementation and evaluation of an automated measurement and visualization method utilizing range point cloud data, *Automation in Construction*, 36, 95-107.

3. Teizer J, Allread BS., Mantripragada U, 2010. Automating the blind spot measurement of construction equipment, *Automation in Construction*, 19(4), 491-501.

4. Teizer J, Golovina O, Wang D, Pradhananga N, 2015. Automated collection, identification, localization, and analysis of worker-related proximity hazard events in heavy construction equipment operation, 2015 Proceedings of the 32st ISARC, Oulu, Finland, 1-9.

5. Ministry of Employment and Labor (2016). The analysis of industrial accident.

6. PSVT, "VIRTUAL 360" <http://www.psvt.com> (Apr. 5, 2017).

7. Chae MJ, Lee GW, Kim JR, Park JW, Yoo HS, Cho MY, 2009. Development of the 3D imaging system and automatic registration algorithm for the intelligent excavation system (IES), *Korean Journal of Construction Engineering and Management*, 4(3), 136-145.

8. Ahn RS, Luo LB, Chong JW, 2010. Real-time implementation of top-view system for camera on vehicle. Proceeding of the Institute of Electronics and Information Engineers Summer Conference, 33(1):1219-1222.

9. Cho YJ, Kim SH, Park JY, Son JW, Lee JR, Kim MH, 2010. Image data loss minimized geometric correction for asymmetric distortion fish-eye lens, *Journal of the Korea Society for Simulation*, 19(1), 23-31.

10. Liu YC, Lin KY, Chen YS, 2008. Bird's-Eye View Vision System for Vehicle Surrounding Monitoring, International Workshop on Robot Vision, 207-218.

11. Continental, "ProViu®ASL360", http://www.continental-automotive.com (Apr. 11, 2017).

12. FUJITSU, "360° Wrap-Around Video Imaging Technology" <http://www.fujitsu.com> (Apr. 5, 2017).

13. ImageNEXT, "OMNIVUE 3D VGA", <http://www.imagenext.co.kr> (Apr. 5, 2017).

14. Yeom DJ, Seo JH, Hong JH, Kim YS, 2015. A conceptual design of around view monitoring system for construction equipment. The 6th International Conference on Construction Engineering and Project Management, 6:662-663.

15. Yeom DJ, Seo JH, Yeom HS, Yoo HS, Kim YS, 2016. The Development of around view monitoring system pilot type for construction equipment. *Korean Journal of Construction Engineering and Management*, 17(3), 143-155.

16. Akyeampong J, Udoka S, Caruso G, Bordegoni M, 2014. Evaluation of hydraulic excavator HumaneMachine Interface concepts using NASA TLX, *International Journal of Industrial Ergonomics*, 44, 374-382.

17. Wang X, Dunston, PS, 2007. Design, strategies, and issues towards an augmented reality-based construction training platform, *Journal of Information Technology in Construction*. 12, 363-380.

18. Langer H, Iversen K, Andersen K, Mouritsen Ø , Hansen R, 2012. Reducing whole-body vibration exposure in backhoe loaders by education of operators, *International Journal of Industrial Ergonomics*, 42 (3), 304-311.

19. Hayn H, Schwarzmann D, 2010. A haptically enhanced operational concept for a hydraulic excavator. In: Zadeh, M.H. (Ed.), Advances in Haptics, pp. 199-220. Vukovar, Croatia.

20. MSHA (Mine Safety and Health Administration), 1999. Haul road inspection handbook (PH99-I-4), U.S. Department of Labor, Washington, DC.

21. Occupational Safety and Health Branch Labour Department, 2005. Code of practice on safe use of excavators, http://www.labour.gov.hk/eng/public/os/B/excavator.pdf (June. 11, 2017).

22. Chi SH, Caldas H, 2012. Image-based safety assessment: automated spatial safety risk identification of earthmoving and surface mining activities, *Journal of Construction Engineering and Management*, 138(3), 341-351.

23. McCollough E, 1893. Photographic topography. Industry: A Monthly Magazine Devoted to Science, Engineering and Mechanic Arts. Industrial Publishing Company, San Francisco, USA, 399-406.

24. Kim JH, Bae JH, Jung WY, 2016. A study on position recognition of bucket tip for excavator, *Journal of Drive and Control*, 13(1), 49-53.

25. Bradley D, Seward D, 1998. The development, control and operation of an autonomous robotic excavator, *Journal of Intelligent and Robotic Systems*, 21, 73-97.

26. Hosseini M, Georganas D, 2003. Design of a multi-sender 3D videoconferencing application over an end system multicast protocol. Proceedings of the eleventh ACM international conference on Multimedia, 480-489.

27. Chen M, 2001. Design of a virtual auditorium, Proceedings of 2001 ACM conference on Multimedia, 19-28.